# Freehand Ultrasound Image Simulation with Spatially-Conditioned Generative Adversarial Networks

Yipeng Hu[1,2], Eli Gibson[1], Li-Lin Lee[3], Weidi Xie[2], Dean C. Barratt[1], Tom Vercauteren[1], J. Alison Noble[2]

[1] Centre for Medical Image Computing, University College London, London, UK
[2] Institute of Biomedical Engineering, University of Oxford, Oxford, UK
[3] Department of Diagnostic Imaging, Royal Free London NHS Foundation Trust, London, UK

**Abstract.** Sonography synthesis has a wide range of applications, including medical procedure simulation, clinical training and multimodality image registration. In this paper, we propose a machine learning approach to simulate ultrasound images at given 3D spatial locations (relative to the patient anatomy), based on conditional generative adversarial networks (GANs). In particular, we introduce a novel neural network architecture that can sample anatomically accurate images conditionally on spatial position of the (real or mock) freehand ultrasound probe. To ensure an effective and efficient spatial information assimilation, the proposed spatially-conditioned GANs take calibrated pixel coordinates in global physical space as conditioning input, and utilise residual network units and shortcuts of conditioning data in the GANs' discriminator and generator, respectively. Using optically tracked B-mode ultrasound images, acquired by an experienced sonographer on a fetus phantom, we demonstrate the feasibility of the proposed method by two sets of quantitative results: distances were calculated between corresponding anatomical landmarks identified in the held-out ultrasound images and the simulated data at the same locations unseen to the networks; a usability study was carried out to distinguish the simulated data from the real images. In summary, we present what we believe are state-of-the-art visually realistic ultrasound images, simulated by the proposed GAN architecture that is stable to train and capable of generating plausibly diverse image samples.

## 1 Introduction

Realistic simulation of medical ultrasound images is at the centre of many computer-assisted medical imaging innovations, such as simulating obstetric examination procedures to facilitate sonographer training [1] and simulating intra-operative images from pre-operative images to enable multimodality image data fusion for surgical planning and guidance [2]. However, one of the ultrasound-specific difficulties in modelling the imaging process is that significant variation during image acquisition comes from unquantified sources of uncertainty.

Physics-based simulation methods for synthesising ultrasound images face significant challenges. Solving large-scale partial differential equations for modelling the nonlinear wave propagation requires substantial computational resource, often infeasible for interactive use. For instance, state-of-the-art simulation needs to make many simplifying assumptions such as ray tracing, static scattering map and requires *a prior* segmentation in order to obtain real-time simulations, e.g. [3]. Furthermore, it is particularly challenging to simulate ultrasound images for individual subjects as acoustic properties of patient-specific soft tissue are in general infeasible to obtain. *In vivo* soft tissue properties are deformable, heterogeneous, and may contain pathologies of interest, all of which vary significantly in individual patients. To the best of our knowledge, there has little published modelling technique available to model acoustic patterns in sonography due to variation at cellular level while modelling motion and deformation at organ level remains on-going research [3].

Manikin based simulators highly depend on the quality of the materials (e.g. realistic acoustic and viscoelastic properties) and the pre-designed fixed anatomy, and hence remains an expensive option. Real patient image databases have also been used in commercial ultrasound simulators. Building these databases relies on nontrivial effort in collecting comprehensive patient cohort, clinical problems and usage cases. There is little published detailed methodology or available data library for research validation. Both methods are therefore limited to pre-defined clinical scenarios for training purposes and are not directly applicable for patient-specific applications such as registering to the pre-operative images of the same patient.

Alternatively, machine learning methods potentially can overcome these restrictions by inferring from real image data. In this work, we are primarily interested in obstetric examination, where freehand ultrasound simulators are increasingly used for training sonographers [1]. Ideal for this clinical scenario, images of the fetus need to be simulated (inferred) at new spatial locations relative to patient anatomy. We argue that fully supervised approaches, such as regression, for predicting ultrasound images are problematic because acquisition-position-independent uncertainties in acquiring training data are both inevitable and significant, such as those caused by the distribution of the acoustic coupling agent, patient motion and other user-dependent variation. For instances, acoustic shadows and refraction artefacts could occur at various regions within the ultrasound fields of view acquired at the same physical location; the speckle pattern changes due to the flow of fluid (e.g. blood) in living tissue. As a result, regression models often lead to blurred averages of nearby training data (see an example in Fig.5) and instantiations that contain sonographic characteristics cannot be easily sampled. Therefore, we propose to use generative adversarial networks (GANs) [4, 5] to model the image distribution as opposed to predicting one single "best" image. From the trained neural network, instances can then be sampled to retain realistic features learned from training data. Furthermore, neural-network-based models can readily generate simulated images on-the-fly without extra engineering effort or specialised hardware.

Using ultrasound image and optical tracking data acquired on a fetus phantom, we summarise our contribution in this proof-of-concept study as follows: a) proposing a novel and stable network architecture for generative modelling of ultrasound images;

b) demonstrating the feasibility of conditional GANs to simulate fetal ultrasound images at locations unseen to the networks; c) quantifying the proposed method using clinically relevant anatomical regions and landmarks; d) producing state-of-the-art visually realistic ultrasound simulation results verified by a usability study.

## 2 Method

### 2.1 Spatially-conditioned generative adversarial learning

In this work, we propose to model the conditional distribution $P_{im}(\mathbf{x}|\mathbf{y})$ of the ultrasound images $\mathbf{x} \sim P_{im}(\mathbf{x}|\mathbf{y})$, given spatial locations $\mathbf{y} \sim P_{loc}(\mathbf{y})$ with respect to a fixed physical reference obtained from a position tracking device. The experiment and data acquisition details are provided in Section 2.3. The ultrasound simulator can then be constructed by optimising the latent parameters $\boldsymbol{\theta}_G$ of a *generator* neural network $G(\mathbf{z}, \mathbf{y})$, so it maps independent unit Gaussian noise $\mathbf{z} \sim N(\mathbf{z})$ to the observed image space for each given spatial location $\mathbf{y}$.

In a zero-sum minimax optimisation framework described in GANs [4], the generator is optimised through the *discriminator* neural network $D(\mathbf{x}, \mathbf{y})$ with latent parameters $\boldsymbol{\theta}_D$, which outputs a scalar likelihood classifying the input as true or false, i.e. real or fake ultrasound image at location $\mathbf{y}$. This is achieved by jointly optimising the cost functions of the discriminator and the generator, $J^{(D)}$ and $J^{(G)}$, respectively:

$$J^{(D)} = -\tfrac{1}{2}\mathbb{E}_{(\mathbf{x},\mathbf{y}) \sim P_{data}} \log D(\mathbf{x}, \mathbf{y}) - \tfrac{1}{2}\mathbb{E}_{\mathbf{z} \sim N, \mathbf{y} \sim P_{loc}} \log(1 - D(G(\mathbf{z}, \mathbf{y}), \mathbf{y})) \quad (1)$$

and

$$J^{(G)} = -\tfrac{1}{2}\mathbb{E}_{\mathbf{z} \sim N, \mathbf{y} \sim P_{loc}} \log D(G(\mathbf{z}, \mathbf{y}), \mathbf{y}) \quad (2)$$

where $\mathbb{E}$ is the statistical expectation. Using sample pairs from data distribution $(\mathbf{x}, \mathbf{y}) \sim P_{data}(\mathbf{x}, \mathbf{y})$, parameters $\boldsymbol{\theta}_D$ and $\boldsymbol{\theta}_G$ are each updated once in every iteration to decrease the values of respective $J^{(D)}$ and $J^{(G)}$ cost functions. Conceptually, optimising $J^{(G)}$ aims to enable $G(\mathbf{z}, \mathbf{y})$ to generate samples that the discriminator classifies as true images; while $J^{(D)}$ is optimised, in an adversarial manner, to "correctly" classify the generator produced images $G(\mathbf{z}, \mathbf{y})$ and samples from the training data set $\mathbf{x}$ as false and true images, respectively. Once convergence is reached, the generator is expected to generate samples from a distribution approximating the conditional data distribution, by only sampling from the $N(\mathbf{z})$ with a given spatial location. The implementation details of the two networks are given in the following sections.

### 2.2 Network architecture

Central to our proposed method is to assimilate the spatial information in an effective and balanced manner, so the non-stationary minimax optimisation can produce images conditioned on the given spatial locations while variations, captured by the input noise, are still preserved. Although, in theory, the physical transformation directly obtained from the tracking could be added to the inputs of the generator and the discriminator for conditioning purpose [4], we found in practice simple concatenation of the transformation vectors in terms of rotation and translation was not effective. The neural networks failed to converge to generating images containing spatially correct anatomy at the given locations. Therefore, we propose: 1) to use calibrated 3D physical coordinates of the image pixels as the input conditioning data (Fig.1 Left); 2) to concatenate the resized x-, y- and z coordinate grids (as three channels) before each up-sampling layer in the generator; 3) to adopt residual network units [6] throughout the discriminator with the conditioning coordinates only being concatenated with input image. An overview of the network architecture is sketched in Fig. 1.

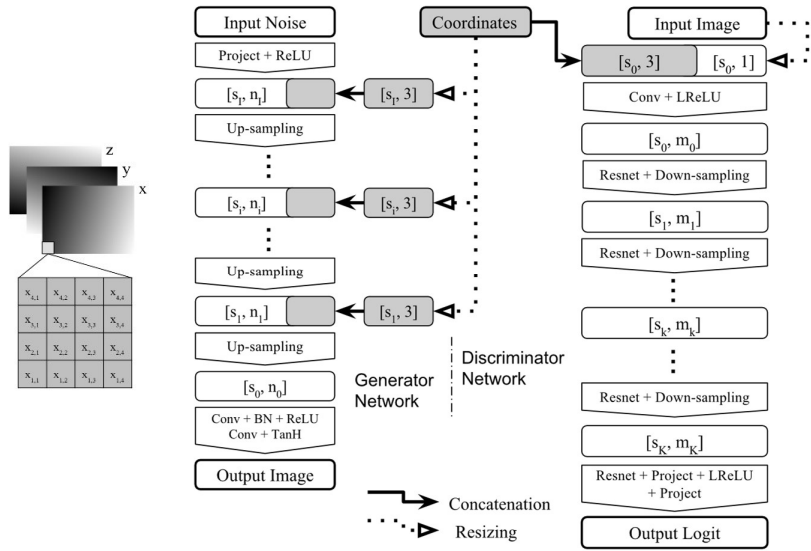

**Fig. 1.** Left: an illustration of the calibrated physical x-, y- and z coordinates of image pixels, contained in three channels; Right: an overview of the neural network architecture used in the proposed conditional generative adversarial networks, where $s_i$ and $s_k$ are sizes of the feature maps while $n_i$ and $m_k$ are numbers of channels. See details in Section 2.2.

In the generator network, 100 random Gaussian noise with zero mean and unit variance is first projected with rectified linear units (ReLU) to feature maps having a small size $s_I$ and $n_I$ initial channels. Then the "up-sampling" layers double the size of the previous feature maps and halve the number of channels, until the size of the image is reached. Each $i^{th}(i = 1, ..., I)$ up-sampling layer consists of a transposed convolution with a 2×2 stride and a convolution (conv), both with batch normalisation (BN) and ReLU. The last layer has a convolution with BN and ReLU, and a second convolution with hyperbolic tangent function (TanH) as activation without BN to retain true data statistics and range [5]. The three conditioning channels containing x-, y- and z pixel coordinates are resized to the respective sizes and concatenated to the feature maps just before each up-sampling (the image-sized coordinate channels were therefore not used directly in the generator). All the convolutional layers in the generator have a 3×3 kernel size.

The discriminator network accepts a concatenation of an input image and its corresponding three-channel pixel coordinates, mapping to feature maps of the same size and $m_0$ initial channels by a convolution with leaky ReLU (LReLU) as activation function, suggested in [5]. Pairs of residual network unit (Resnet) and "down-sampling" layer halve the size of the previous feature maps and double the number of channels. Each $k^{th}(k = 1, ..., K)$ Resnet layer has two convolutions both with BN and LReLU, and an identity mapping for shortcut. Each down-sampling layer has a 2×2 stride convolution with BN and LReLU. The final Logit with one-sided label smoothing [7] is outputted after an additional Resnet, two projections (to a single image size and to a scalar, respectively) with a nonlinear LReLU in between. All the convolutional layers in the discriminator also have a 3×3 kernel size, with an exception of the first one having a larger 5×5 kernel.

### 2.3 Validation experiment

An approximately one hour scan of an anatomically realistic fetus ultrasound examination phantom ("SPACE FAN-ST", Kyoto Kagaku Co., Ltd, Kyoto, Japan) was performed by a Reporting Sonographer with more than ten years' experience. As illustrated in Fig.2, the abdominal probe (Ultrasonix 4DC7-3/40, BK Ultrasound, BC, Canada) was tracked by an optical tracker (Polaris Spectra, NDI Europe, Radolfzell, Germany). The ultrasound images and tracking data were timestamped by NifTK (niftk.org) [8]. The data were acquired in four sessions at neurological, cardiological, abdominal and fetal profile regions, following standard NHS Fetal Anomaly Screening Programme.

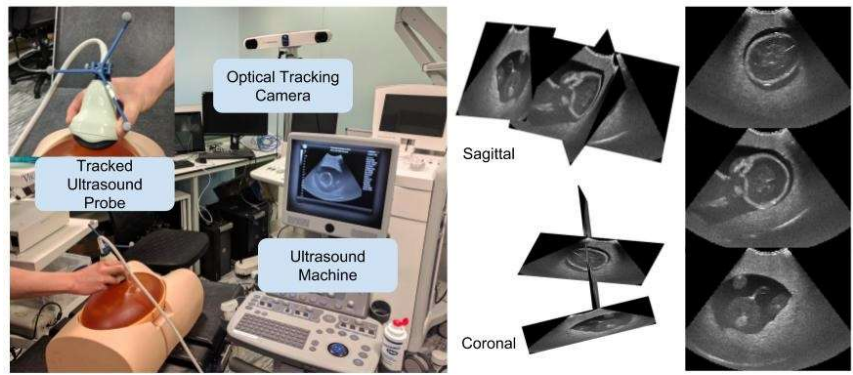

**Fig. 2.** Left: the laboratory setup for acquiring the ultrasound image and optical tracking data; Right: example ultrasound frames plotted both in 3d (using calibrated tracking data) and in 2d.

A total of 26,396 image frames were normalised to an intensity range of $[-1,1]$ and were down-sampled to $160\times120$ pixels in this study, only images containing anatomical structures being used in this study for computational consideration. The physical pixel coordinates were calibrated based on a pinhead-based calibration method [9], before being normalised to zero-mean and unit variance. The calibration procedure acquired 87 tracked 2D ultrasound images of a fixed pinhead at different angles and positions to estimate the relative transformation from local image coordinates to global physical coordinates together with the in-plane resolution.

The proposed GANs-based ultrasound simulator was implemented with Tensor-Flow™ and trained on a 12GB NVIDIA® Tesla™ K40 GPU, using the Adam optimiser with learning rate set to 0.0002, first- and second moment estimates set to 0.5 and 0.999, respectively. The results presented in this paper was obtained with a minibatch size of 36, no weight decay or clipping, $I=4$ down-sampling- and $K=5$ up-sampling layers, 512 and 32 initial channels for the generator and the discriminator, respectively.

To assess the anatomical fidelity of the simulation, clinically interesting landmarks, including crux of four heart chambers, centre of stomach, cord insertion, mid-line echo, cavum septum pellucidum and nasal tip on profile, were identified in both the held-out real images and the simulated images generated from a 10-fold cross-validation experiment (see examples in Fig.3). Images were also excluded from training data if the tracked transformation (in rotation and translation) is within 95% confidence interval [10] of any transformation in those of held-out test data.

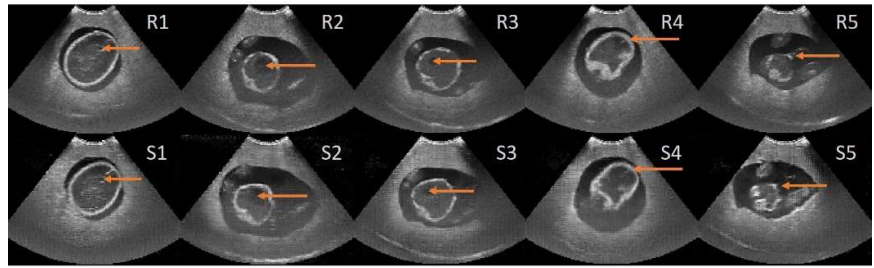

**Fig. 3.** Example anatomical landmarks (orange arrows) used in this work, between real images (top row, R1-5) and simulated images sampled at the same locations (bottom row, S1-5).

For comparison with regression-like approaches, two additional models were also trained: 1) a heavily-supervised GANs with an equally-weighted $L^2$-norm regularisation term [11] added to the generator's cost function (Eq. 2), with other settings unchanged; 2) a regression model of the same generator architecture, directly minimising $L^2$-norm of the difference between generated- and training images.

A usability study was conducted, in which the sonographer was asked to distinguish whether the images are simulated or real. Randomly sampled 100 generated simulations together with 100 real ultrasound images were shuffled, before being displayed full-screen on a 15-inch monitor. To further quantify the GANs' ability to generate realistic content at different spatial frequencies, the experiment was repeated while a Gaussian filter was applied with different kernel sizes (standard deviation σ ranged from 2 to 0 mm, with zero indicating that the original images produced by the network and the original real images were used directly).

## 3    Results

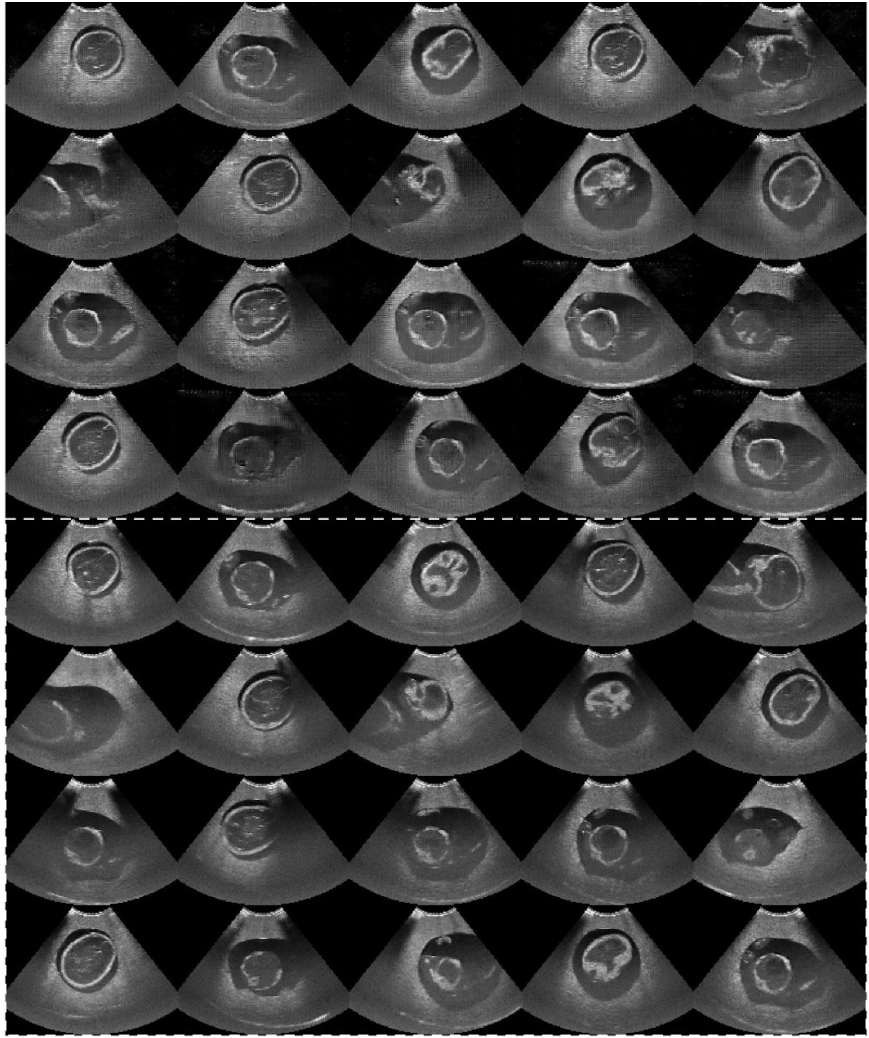

**Fig. 4.** Top four rows: examples of randomly-sampled ultrasound image simulations at unseen spatial locations; bottom four rows: real ultrasound images at the same locations in the validation data set.

The generator network can produce more than 1k frames per second (fps) on the same GPU and ~60 fps on an Intel® Core™ i7 CPU, well satisfying real-time requirement. The examples of simulated ultrasound images at locations unseen to the networks are provided in Fig.4, together with corresponding ground-truth images at the same spatial locations. Verified by the same sonographer who acquired the data, 120 landmarks were identified in the held-out real images. Among the randomly sampled 120 simulated images, 3 (2.5%) were considered as producing incorrect or unrecognisable regions, while 47 (39.2%) simulated images contain recognisable anatomical regions but no clearly identifiable corresponding landmarks. From the remaining 70 (58.3%) image pairs, 2D distances were calculated between the corresponding landmarks from simulated- and ground-truth images, yielding a mean 6.1±1.2 mm error.

Fig.5 contains example simulated images from 1) the GANs trained with heavily supervised regularisation and 2) the regression model, both described in Section 2.3. It demonstrates that, compared with the images from the proposed simulator, the apparent blurring effect is predominant from the two more supervised learning approaches with inferior visual features and details.

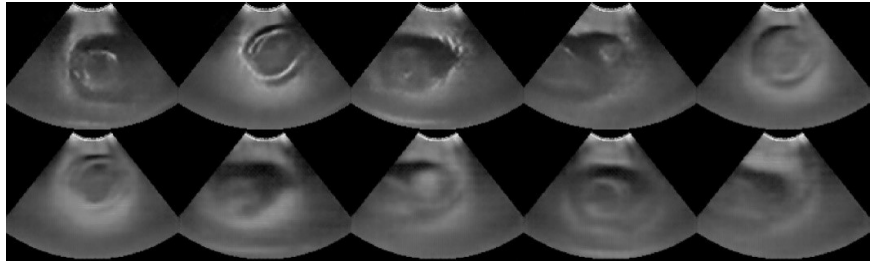

**Fig. 5.** Examples of simulations using the supervised cost functions: (1) Top row: the images sampled from the trained GANs with heavily supervised regularisation; (2) Bottom row: the regression predicted images with the $L^2$-norm objective function (see details in Section 2.3).

We report an interesting result from the usability study described in Section 2.3. The sonographer was able to correctly distinguish 124 (62%) and 162 (81%) test images, with Gaussian filtering (σ=1 mm) and without filtering (σ=0 mm), respectively. The difference may partly be explained by the network-related high frequency artefacts (e.g. checkboard artefacts in transposed convolution [12]) being filtered with larger kernels. We note that any Gaussian kernel with a size larger than σ=1.5 mm was deemed too blur to avoid random guess (approaching 50% correction rate).

To further investigate the variance learned by the generative models, Fig.6 illustrates the simulated images by only sampling the prior noise **z** with fixed conditioning **y**. It shows that, also found in Fig.4, changes in detailed intensity patterns, positions of shadows and artefacts, and minor anatomical variation may be captured by modelling the conditional distribution of ultrasound images.

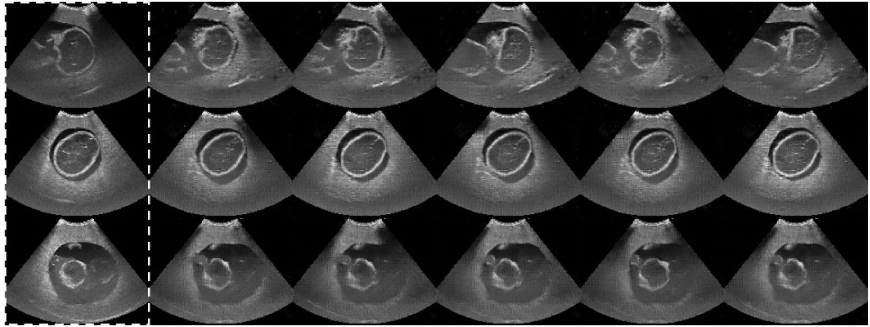

**Fig. 6.** Examples of randomly-sampled ultrasound image simulations at the same spatial locations (rows) with the first column displaying the ground-truth real ultrasound images acquired at the same locations.

## 4   Conclusion and Discussion

Based on both the qualitative and quantitative results reported in this paper, we present a promising approach for generative representation learning of freehand ultrasound images in fetal examination.

Our experience suggests that the proposed GAN architecture, together with the calibrated physical coordinates as conditioning input, not only reduces the training time but also stabilises the convergence, e.g. the training was relatively insensitive to hyperparameters and no divergence or imbalanced cost functions were observed in the experiments. This is probably a result of the presumably simplified training objective and better supervision from the conditioning information at different resolution levels. However, further results are needed to draw a comprehensive conclusion.

There has been evidence of *mode-collapse* at some physical locations. Arguably, it may not be critical in this application that aims to produce high quality samples instead of complete coverage of the image distribution. A full investigation of this well-recognised problem is beyond the scope of the current work, and remains an open research question, e.g. [7]. Minor *underfitting* has been observed with the presented GANs, evidenced by the fact that the samples drawn from locations in the training set are very similar (both in subjective appearance and quantitative landmark distance) to those found in testing. Therefore, we believe that, considering the variation in the learned conditional distribution, the reported landmark error reflects the modelling ability of generating spatially coherent anatomical features, rather than an indication of copying nearest training data.

Although deformation was commonplace during the experiment due to probe pressure and the soft mattress of the surgical bed on which the deformable phantom was placed, future research will aim to apply the method on real patient data which exhibit more complex variation such as challenging fetal movement. A wider range of training and test data (e.g. acquired by non-experts) may need to be included for the purpose of training less experienced users, and investigating other types of conditioning information (e.g. ultrasound parameters and temporal variation) could also improve the interactive capacity of the proposed simulator. We would also like to note that, although an optical tracker was used in this validation, the trained simulator may be feasible to run on consumer-grade mobile devices equipped with inexpensive inertial measurement units.

## Acknowledgement

The authors would like to thank Dr Matt Clarkson, Dr Steve Thompson, and Dr Francisco Vasconcelos from UCL for assistance with tracked ultrasound data collection on the fetus phantom. The source code used in this work is available on NiftyNet platform (niftynet.io) under an Apache 2.0 license.

## Reference

1. Maul, H., et al. (2004). Ultrasound simulators: experience with the SonoTrainer and comparative review of other training systems. Ultrasound in obstetrics & gynecology, 24(5), 581-585.
2. Wein, W., et al. (2007). Simulation and fully automatic multimodal registration of medical ultrasound. Medical Image Computing and Computer-Assisted Intervention–MICCAI 2007, 136-143.
3. Burger, B., et al. (2013). Real-time GPU-based ultrasound simulation using deformable mesh models. IEEE transactions on medical imaging, 32(3), 609-618.
4. Goodfellow, I., et al. (2014). Generative adversarial nets. In Advances in neural information processing systems (pp. 2672-2680).
5. Radford, A., Metz, L., & Chintala, S. (2015). Unsupervised representation learning with deep convolutional generative adversarial networks. arXiv preprint arXiv:1511.06434.
6. He, K., et al. (2016). Deep residual learning for image recognition. In Proceedings of the IEEE Conference on Computer Vision and Pattern Recognition (pp. 770-778).
7. Salimans, T., et al. (2016). Improved techniques for training gans. In Advances in Neural Information Processing Systems (pp. 2234-2242).
8. Clarkson, M. J., et al. (2015). The NifTK software platform for image-guided interventions: platform overview and NiftyLink messaging. International journal of computer assisted radiology and surgery, 10(3), 301-316.
9. Hu, Y., et al. (2017). Development and Phantom Validation of a 3-D-Ultrasound-Guided System for Targeting MRI-Visible Lesions During Transrectal Prostate Biopsy. IEEE transactions on bio-medical engineering, 64(4), 946.
10. Elfring, R., de la Fuente, M., & Radermacher, K. (2010). Assessment of optical localizer accuracy for computer aided surgery systems. Computer Aided Surgery, 15(1-3), 1-12.
11. Pathak, et al. (2016). Context encoders: Feature learning by inpainting. In Proceedings of the IEEE Conference on Computer Vision and Pattern Recognition (pp. 2536-2544).
12. Odena, A., Dumoulin, V., & Olah, C. (2016). Deconvolution and checkerboard artifacts. *Distill*, *1*(10), e3.